\documentclass[letterpaper, 10 pt, conference]{ieeeconf}  % Comment this line out if you need a4paper

\IEEEoverridecommandlockouts                              % This command is only needed if 
\title{\LARGE \bf
Learning Adaptive Dexterous Grasping from Single Demonstrations
}

\author{
Liangzhi Shi$^{1,2}$\authorrefmark{1}, Yulin Liu$^{1}$\authorrefmark{1}, Lingqi Zeng$^{1}$\authorrefmark{1}, Bo Ai$^{1}$, Zhengdong Hong$^{1}$, and Hao Su$^{1,3}$\\
$^{1}$University of California, San Diego \quad $^{2}$Tsinghua University  \quad $^{3}$Hillbot\vspace{-5pt}
\thanks{\authorrefmark{1} Equal contribution.}
}

\usepackage{cite}
\usepackage{amsfonts}
\let\labelindent\relax
\usepackage{enumitem}  
\usepackage{float}
\usepackage{afterpage}
\usepackage{stfloats}
\usepackage{tabularx}
\usepackage{array}
\usepackage{makecell}
\usepackage{graphicx}
\usepackage{booktabs}
\usepackage{multirow}
\usepackage{dsfont}
\usepackage{romannum}
\usepackage{hyperref}
\usepackage[dvipsnames]{xcolor}
\usepackage{xspace}
\usepackage{tabularray}
\usepackage{amsmath} % assumes amsmath package installed
\usepackage{amssymb}  % assumes amsmath package installed
\usepackage{subcaption}
\usepackage{tabularx}

\newcommand{\secref}[1]{Section~\ref{#1}}
\renewcommand{\eqref}[1]{Eqn~\ref{#1}}
\newcommand{\figref}[1]{Figure~\ref{#1}}

\newcommand{\tabref}[1]{Table~\ref{#1}}

\newcommand{\eg}{\textrm{e.g.}}

\def\ours{AdaDexGrasp\xspace}

\def\mustard{\texttt{mustard-bottle}}
\def\bleach{\texttt{bleach-cleanser}}
\def\cracker{\texttt{cracker-box}}
\def\sugar{\texttt{sugar-box}}

\begin{document}

\maketitle
\thispagestyle{empty}
\pagestyle{empty}

\newcommand{\update}[1]{{#1}}
\newcommand{\short}[1]{{#1}}

%%%%%%%%%%%%%%%%%%%%%%%%%%%%%%%%%%%%%%%%%%%%%%%%%%%%%%%%%%%%%%%%%%%%%%%%%%%%%%%%
\begin{abstract}

How can robots learn dexterous grasping skills efficiently and apply them adaptively based on user instructions? This work tackles two key challenges: efficient skill acquisition from limited human demonstrations and context-driven skill selection. We introduce AdaDexGrasp, a framework that learns a library of grasping skills from a single human demonstration per skill and selects the most suitable one using a vision-language model (VLM). To improve sample efficiency, we propose a trajectory following reward that guides reinforcement learning (RL) toward states close to a human demonstration while allowing flexibility in exploration. To learn beyond the single demonstration, we employ curriculum learning, progressively increasing object pose variations to enhance robustness. At deployment, a VLM retrieves the appropriate skill based on user instructions, bridging low-level learned skills with high-level intent. We evaluate AdaDexGrasp in both simulation and real-world settings, showing that our approach significantly improves RL efficiency and enables learning human-like grasp strategies across varied object configurations. Finally, we demonstrate zero-shot transfer of our learned policies to a real-world PSYONIC Ability Hand, with a 90\% success rate across objects, significantly outperforming the baseline.

% keywords: human-like dexterous grasping and preference-aware skill selection

% contribution: We propose a framework for preference-aware dexterous grasping that learns a library of human-like skills from single demonstrations and leverages a vision-language model for skill selection based on user preferences.

\end{abstract}
%%%%%%%%%%%%%%%%%%%%%%%%%%%%%%%%%%%%%%%%%%%%%%%%%%%%%%%%%%%%%%%%%%%%%%%%%%%%%%%%

\section{Introduction}

Imagine handing a power drill to a teammate during assembly. We naturally grasp it from the top for an easy handoff, but grip the handle when using it ourselves. Humans instinctively adjust their grasp to the task, while robots often rely on fixed strategies. Even learning a single dexterous grasp is challenging due to the hand's high degrees of freedom. How can robots acquire varied grasping skills and apply them adaptively?

% Addressing this requires tackling two key challenges: (i) efficiently learning dexterous grasping skills and (ii) adaptively selecting skills based on context. 

The first challenge is to acquire dexterous grasping skills efficiently. Classical robotic control relies on perfect state information, which is often impractical due to perception challenges \cite{9815144,rus1999hand, mordatch2012contact, wu2022learning}. Learning-based approaches attempt to mitigate this by leveraging human demonstrations through imitation learning, which requires extensive data collection and specialized hardware \cite{brohan2022rt,qin2023anyteleop, ding2024bunny, wang2024dexcap}, or reinforcement learning (RL), which suffers from low sample efficiency due to the large state and action space \cite{rajeswaran2017learning,chen2024vividex}. In this work, we hypothesize that the structural similarity between human and robot hands makes human data a useful \textit{prior}, not perfect demonstrations, for robotic dexterous grasping. While a small number of demonstrations may effectively guide exploration in RL, the optimal way to leverage imperfect demonstrations remains an open question.

% To this end, we incorporate limited human trajectories to guide exploration in RL, improving sample efficiency and enabling the acquisition of human-like grasps without being constrained by imperfections in the demonstrations.

% Given the similarity between dexterous and human hands, human demonstrations offer valuable guidance for learning grasping policies. 
%However, existing methods for incorporating human data into RL remain inefficient. 
% Therefore, we seek to learn a library of dexterous grasping skills, each derived from a single human-object interaction trajectory, using a trajectory-following reinforcement learning approach.

Grounding robot actions in context is another challenge. Vision-language action models (VLAs) attempt to map language and images directly to robot actions but require massive datasets and typically focus on kinematically simple end-effectors, such as parallel grippers \cite{kim24openvla, octo_2023, rt22023arxiv, liu2024rdt}. Inspired by prior work leveraging vision-language models for language-guided task planning \cite{yang2024guiding, 10611112, wu2023integrating, saycan2022arxiv, codeaspolicies2022}, we aim to integrate learned grasping skills with a language-driven selection mechanism to enable adaptive grasping.  

To this end, we introduce \ours{}, a framework with three key components. First, we propose a trajectory following that encourages an RL agent to visit states close to a reference trajectory, which is obtained from a human demonstration via retargeting. This improves sample efficiency while preserving human-like grasping strategies. Second, we introduce a curriculum learning scheme that gradually randomizes the initial object pose, allowing the policy to generalize beyond the single demonstration. Finally, we construct a skill library using the learned policies and query a vision-language model to select the most appropriate skill based on user preferences at deployment. 

We extensively evaluate \ours{} in both simulation and the real world. Our results show that the proposed reward function alone significantly improves sample efficiency, raising the success rate from near zero to an average of 64\%. In addition, the proposed curriculum learning strategy further enables the policy to handle varied object configurations not present in human demonstrations. To bridge high-level instructions with low-level execution, our vision-language model achieves a 90\% success rate in selecting the appropriate grasping skill based on user input. Finally, we deploy our policy on a real-world PSYONIC Ability Hand, achieving a success rate above 90\% consistently, while the baseline struggles to grasp the object successfully.

% human demo + retargeting -> robot demo + IL
% vividex:  

\section{Related Work}

\subsection{Dexterous Manipulation}

% \bo{double check references. we need a few more representative papers for each of these categories: classical planning, IL and RL (if this is the right categorization, based on what we have now)}

Dexterous manipulation is highly challenging due to the high degree of freedom of the end effector. 
% Different from other end-effectors like the parallel gripper, dexterous manipulation uses high-dimensional end-effectors like anthropological hands which are much more complex in degrees of freedom, posing large challenges in practice.
Traditional methods that use trajectory optimization \cite{rus1999hand, mordatch2012contact, wu2022learning, he2025learning} require precise modeling of robot dynamics, which is difficult in the open world. In contrast, learning-based methods seek to obtain an observation-to-action mapping from data. 
% Some recent works focus on learning-based methods to solve this problem. They try to solve the task using reinforcement learning (RL) \cite{chen2024object, wang2024cyberdemo, qi2023general, singh2024hand} or imitation learning (IL) \cite{qin2022dexmv, chen2024vividex}. 
Reinforcement learning methods \cite{qi2023general, yuan2024robot, wang2024lessons, honglearning} often require extensive exploration in simulation and specific reward engineering. In addition, the learned grasping policies from reward engineering could be unnatural when compared with human. %, but the sim-to-real gap is a significant challenge. 
Imitation learning methods \cite{bain1995framework, chi2023diffusion} are able to leverage real-world data collected with teleoperation systems \cite{qin2023anyteleop, ding2024bunny, wang2024dexcap}, but real-robot data is costly to acquire and the policy is limited by the quality and amount of expert demonstrations. In this work, we aim to leverage a single real-world \textit{human} demonstration to accelerate reinforcement learning, alleviate the burden for reward engineering, and acquire human-like grasp poses while using a learning curriculum to enable generalization beyond the demonstrated object configuration.

\subsection{Learning from Human-Object Interaction Data}

\short{Human-object interaction datasets with 3D human pose and 6-DoF object pose annotations (\eg, \cite{chao2021dexycb, fan2023arctic, liu2022hoi4d}) provide rich resources for learning dexterous manipulation. \cite{qin2022dexmv} extracts robot and object poses from human videos for imitation learning, but suffers from noisy state-action pairs and relies on hundreds of demonstrations and extensive reward tuning. To overcome these issues, \cite{chen2024vividex} proposes a trajectory mapping reward for one-shot learning, using trajectory augmentation and policy distillation to improve generalization. However, this reward is suboptimal as it enforces alignment with potentially flawed demonstrations. Other methods generate human-object trajectories as RL references \cite{chen2024object}, or refine imperfect policies learned from retargeted demonstrations via RL or imitation \cite{singh2024hand}. In contrast, we learn a grasping skill library using a trajectory following reward and curriculum learning, enabling sample-efficient, human-like grasp acquisition and adaptive skill selection.}
% Human-object interaction datasets labelled with 3D human pose and 6-DoF object pose (\eg, \cite{chao2021dexycb, fan2023arctic, liu2022hoi4d}) are a rich repository for learning dexterous manipulation skills.  \cite{qin2022dexmv} extracts robot and object poses from human videos for imitation learning but suffers from noisy state-action pairs, requiring hundreds of demonstrations and extensive reward engineering. To address these limitations, \cite{chen2024vividex} introduces a trajectory mapping reward to learn from a single demonstration, leveraging trajectory augmentation and policy distillation for better generalization. However, its trajectory mapping reward is suboptimal as it enforces alignment with potentially imperfect human demonstrations. Other approaches learn to generate human-object interaction trajectories as reference trajectories in RL \cite{chen2024object} or learn an imperfect policy from retargeted human demonstrations, refining it via RL or imitation learning \cite{singh2024hand}. Our work differs by learning a library of grasping skills using a trajectory following reward and a learning curriculum, enabling sample-efficient acquisition of human-like grasps and adaptive skill selection.  

\subsection{Language-Grounded Manipulation}

Aligning robot behaviors with human preference is an important research direction  \cite{wang2024inference, rth2024arxiv}.
% Beyond solving the problem of dexterous manipulation, an important research direction focuses on human-machine interaction and aligning robotic behavior with human preferences  \cite{wang2024inference, rth2024arxiv}. 
Language instructions are widely used to convey human preferences \cite{luketina2019survey, jang2022bc, stepputtis2020language, shridhar2022cliport, mees2022matters}, and prior work \cite{rt22023arxiv, octo_2023, kim24openvla, liu2024rdt} has developed Vision-Language-Action (VLA) models to enable robots to follow language commands. However, these models primarily focus on high-level task specifications rather than low-level control variations, such as different grasp poses.

\short{An alternative approach to incorporating human preferences leverages Vision-Language Models (VLMs) \cite{gao2025vision, bommasani2021opportunities} with structured skill libraries \cite{saycan2022arxiv,zha2023distilling, codeaspolicies2022, wu2023integrating, wu2025savor}. These methods use VLMs to interpret language instructions, extract preference-related information, and retrieve suitable skills from a predefined set. While effective in various manipulation tasks, their application to dexterous manipulation remains limited, particularly in adapting grasp poses to user intent. This work extends the paradigm by building a skill library from human demonstrations and integrating a VLM to enable preference-driven grasp selection, bridging the gap between language understanding and dexterous control.}
% An alternative approach to incorporating human preferences leverages Vision-Language Models (VLMs) \cite{gao2025vision, bommasani2021opportunities} alongside structured skill libraries \cite{saycan2022arxiv,zha2023distilling, wu2023integrating, codeaspolicies2022, wu2025savor}. These methods use VLMs to interpret language instructions, extract preference-related information, and retrieve suitable skills from a predefined set. While effective in various manipulation tasks, their application to dexterous manipulation remains limited, particularly in adapting grasp poses based on user intent. This work extends this paradigm by constructing a skill library from human demonstrations and integrating a VLM to enable preference-driven grasp selection, bridging the gap between language understanding and dexterous control.

\section{Method} \label{sec:method}

\subsection{Overview}

The objective of this work is to enable human-like, preference-aware dexterous grasping. To achieve this, the robot must be able to acquire diverse grasping strategies and select the most appropriate one based on human instructions.

We approach this problem by first learning a library of grasping skills from human demonstrations and then developing a mechanism to select a skill based on environment observations and natural language commands. Each skill in the library is trained via reinforcement learning with trajectory guidance, improving the sampling efficiency of learning while ensuring that the learned grasping behaviors resemble human demonstrations. Given an environment observation, a natural language instruction, and a set of skill descriptions, a selection module determines the best skill for execution.

Formally, we define the problem as a Markov Decision Process (MDP) $\mathcal{M} = (\mathcal{S}, \mathcal{A}, P, r, \gamma)$. At each time step $t$, the robot observes the state $s_t \in \mathcal{S}$ and selects an action $a_t$ according to a skill policy $\pi$:
\begin{equation*}
    a_t \sim \pi(a_t \mid s_t).
\end{equation*}
The environment then transitss to a new state $s_{t+1}$ according to the dynamics model $P$, and the robot receives a reward $r_t$. Each skill policy $\pi_i \in \Pi$ is optimized to maximize the expected cumulative reward given a reference trajectory extracted from a human demonstration $\tau_i$:
\begin{equation*}
    \pi^* = \arg\max_{\pi} \mathbb{E} \left[ \sum_{t=0}^{H} \gamma^t r(s_t, a_t, s_{t+1} \mid \tau_i) \right],
\end{equation*}
\short{where $H$ is the task horizon. The reference $\tau_i$ (e.g., a human grasping video) both guides learning for sample efficiency and promotes human-like behavior.}
% where $H$ is the task horizon, and $\tau_i$ is a reference trajectory, which could be a video of a human grasping an object. The reference trajectory serves two purposes: (i) guiding reinforcement learning to improve sample efficiency, and (ii) encouraging the learned grasping strategy to align with human behavior.

Once the skill library is learned, we introduce a skill selection function $\Psi$. Given an environment image $P$, a natural language instruction $I$, and a set of skills with language descriptions $\Pi$, $\Psi$ retrieves the most suitable skill for execution:
\begin{equation*}
    \pi^* = \Psi(P, I, \Pi).
\end{equation*}

This framework allows the robot to acquire grasping strategies from single human demonstrations and adapt its grasp based on preferences in human instruction. Next, we detail how we extract human demonstrations (\secref{sec:demo}), train policies with trajectory guidance (\secref{sec:rl}) via curriculum learning (\secref{sec:cl}), and how the selection module enables preference-aware grasping (\secref{sec:skill}).

% (1) human video to demo
% (2) Curriculum Learning
% (3) VLM + skill inventory, grasp everything

% This section outlines the details of our proposed method, which is illustrated in Figure X. Our approach consists of three key components: (1) Converting human manipulation videos into robot hand manipulation demonstrations. (2) Curriculum Learning: Enabling the model to learn object grasping from demonstrations across various poses. (3) Building the agent using a Vision-Language Model (VLM).

\begin{figure*}
    \centering
    \includegraphics[width=\linewidth]{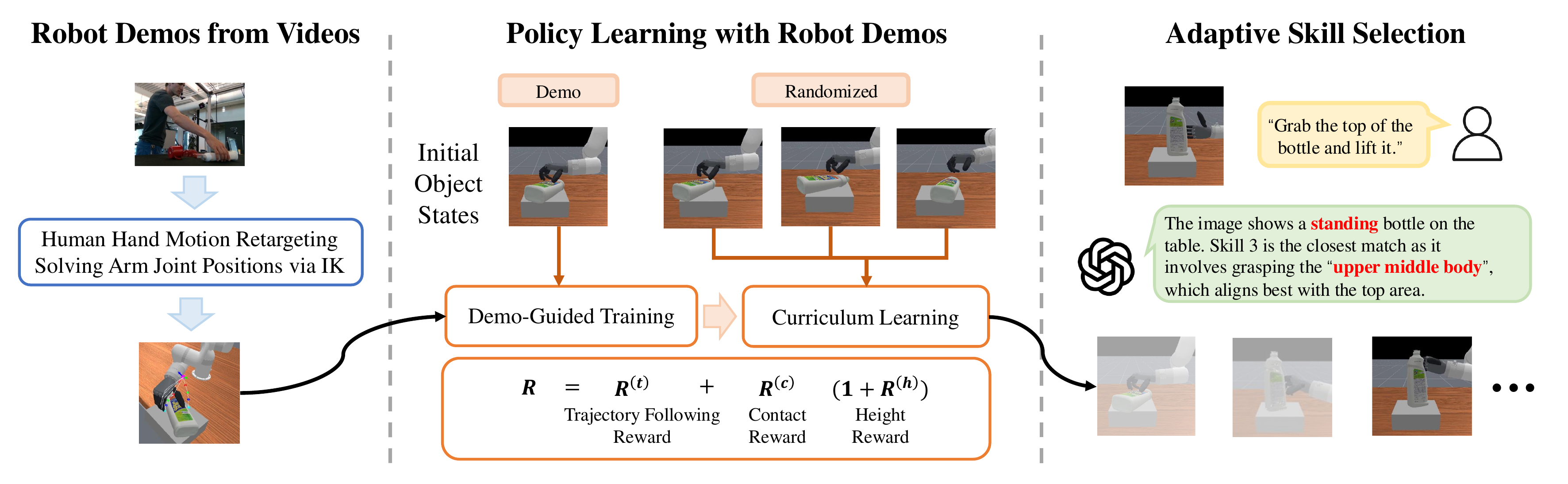}\vspace{-10pt}
    \caption{Method overview. \ours{} consists of three key components: (i) Converting human videos into robot demonstrations via motion retargeting, (ii) learning a grasp policy through trajectory-guided reinforcement learning with a curriculum, and (iii) skill selection using vision-language models to ground robot behavior into user preference.}
    \label{fig:main_pipeline} \vspace{-20pt}
\end{figure*}

\subsection{Human Videos as Robot Demonstrations} \label{sec:demo}

% We use hand and object poses from the DexYCB\cite{chao2021dexycb} dataset, a real-world video datasets that contain human demonstrations of hand-object interactions. DexYCB datasets present a variety of objects, each typically associated with at least two initial positions and two grasping postures.

% We use the DexYCB dataset, which provides real-world videos with ground-truth hand and object poses for various items, such as a sugar box and power drill \cite{chao2021dexycb}. Each object has demonstrations with at least two initial positions and grasp poses. Our goal is to convert these human demonstrations into robot trajectories for reinforcement learning.

% To bridge the gap between human data and robot hardware, we first map human hand motions to the robot’s hand joint positions. Since DexYCB hand poses are not attached to an arm, we anchor the robot’s hand to the arm for realistic motion. The process involves (i) retargeting the hand motion to the robot's joints and (ii) using inverse kinematics (IK) to solve the arm joint positions. This allows the robot to replicate human grasping while respecting kinematic constraints.

\short{To obtain human demonstrations, we use the DexYCB dataset, which provides real-world video recordings along with ground-truth hand and object poses for various objects\cite{chao2021dexycb}. Each object has demonstrations with at least two initial positions and two grasp poses. We seek to translate human demonstrations to robot trajectories to guide downstream reinforcement learning. }
% To obtain human demonstrations, we use the DexYCB dataset, which provides real-world video recordings along with ground-truth hand and object poses for various objects, such as a sugar box, mustard bottle, and bleach cleanser \cite{chao2021dexycb}. For each object, the dataset includes demonstrations with at least two initial object positions and two distinct grasp poses. We seek to translate human demonstrations to robot trajectories to guide downstream reinforcement learning. 

The challenge lies in overcoming the embodiment gap between human data and robot hardware. To this end, we first translate human hand motions into robot joint positions to ensure kinematic feasibility with the robotic system. Since the extracted hand poses from DexYCB are floating and not attached to an arm, we explicitly anchor the robot hand to an arm to maintain realistic motion representation. The translation process consists of two steps: (i) retargeting hand motion, where human hand motions are mapped to the robot’s hand joint positions, and (ii) solving joint positions of the arm with inverse kinematics (IK). This process enables the robot to replicate human grasping strategies while maintaining kinematic constraints.

\textbf{Hand motion retargeting.} 
Following \cite{qin2023anyteleop, ding2024bunny, chen2024vividex, handa2020dexpilot,antotsiou2018task}, we 
% assume a floating robot hand in the initial step and 
adopt an optimization-based approach. 
% to hand motion regargeting through an optimization-based view hand motion retargeting as an optimization problem. 
The goal is to minimize the difference between the fingertip positions of the robot hand and those of the human hand. The loss function is: 
% \begin{gather}
%     \min_{q_t}\sum_{i=0}^{N} \Vert x_t^i - f_i(q_t) \Vert^2 + \beta \Vert q_t - q_{t-1} \Vert^2 \\
%     \text{s.t. } \quad q_l \leq q_t \leq q_u
% \end{gather}
\begin{equation*}
    \mathcal{L}_{\text{retarget}}(q_k) = \sum_{i=1}^{5} \Vert x_k^i - f_i(q_k) \Vert^2 + \beta \Vert q_k - q_{k-1} \Vert^2,
\end{equation*}

subject to:
\begin{equation*}
    q_l \leq q_k \leq q_u,
\end{equation*}
where $q_k$ denotes the robot hand joint positions at time $k$, $x_k^i$ is the $i$-th fingertip position of the human hand, $f_i(q_k)$ represents the corresponding $i$-th fingertip of robot hand given $q_k$ via forward kinematics, and $q_l$ and $q_u$ are the lower and upper bound limits of the joint positions. An additional penalty term weighted by $\beta$ enhances temporal smoothness. We perform optimization using the NLopt solver \cite{qin2023anyteleop}. %\yulin{need to explain more about mimic joint}.

% \textbf{Computing arm joints via inverse kinematics.} 
% To ensure physically plausible movements, We attach the floating robot hand to a robot arm and employ Closed-loop Inverse Kinematics (CLIK), implemented with the Pinocchio library \cite{pinocchioweb, carpentier2019pinocchio}, to calculate the joint positions of the robot arm. The target pose for this IK computation is the end-effector pose of the initially assumed floating hand. This step verifies feasibility of hand configurations, ensuring the robot hand motions remain consistent with real-world constraints. 

% Shown by physiological study, human hand motion follows a minimum jerk trajectory\cite{flash1985coordination}, where jerk is defined as the absolute value of the third-order derivative of motion trajectories. Minimizing jerk helps reduce joint position errors in motors \cite{kyriakopoulos1988minimum} and limits excessive wear on the physical robot \cite{merat1987introduction}. Based on this principle, we apply post-processing to the sequence of joint positions $\{q_t\}_{t=0}^{T}$ following \cite{qin2022dexmv}. Specifically, we apply the model proposed in \cite{todorov1998smoothness}. Fig. \ref{fig:retargeting} illustrates an example of motion retargeting for ability hand attached to an Xarm7 arm. 

\textbf{Computing arm joints via IK.}  Given end-effector poses from the retargeted robot hand trajectory, we compute the joint positions of the arm using Closed-Loop Inverse Kinematics (CLIK) with the Pinocchio library \cite{pinocchioweb, carpentier2019pinocchio}. 
%The target pose for IK is the end-effector pose of the floating hand. 

\begin{figure}
    \centering
    \includegraphics[width=\linewidth]{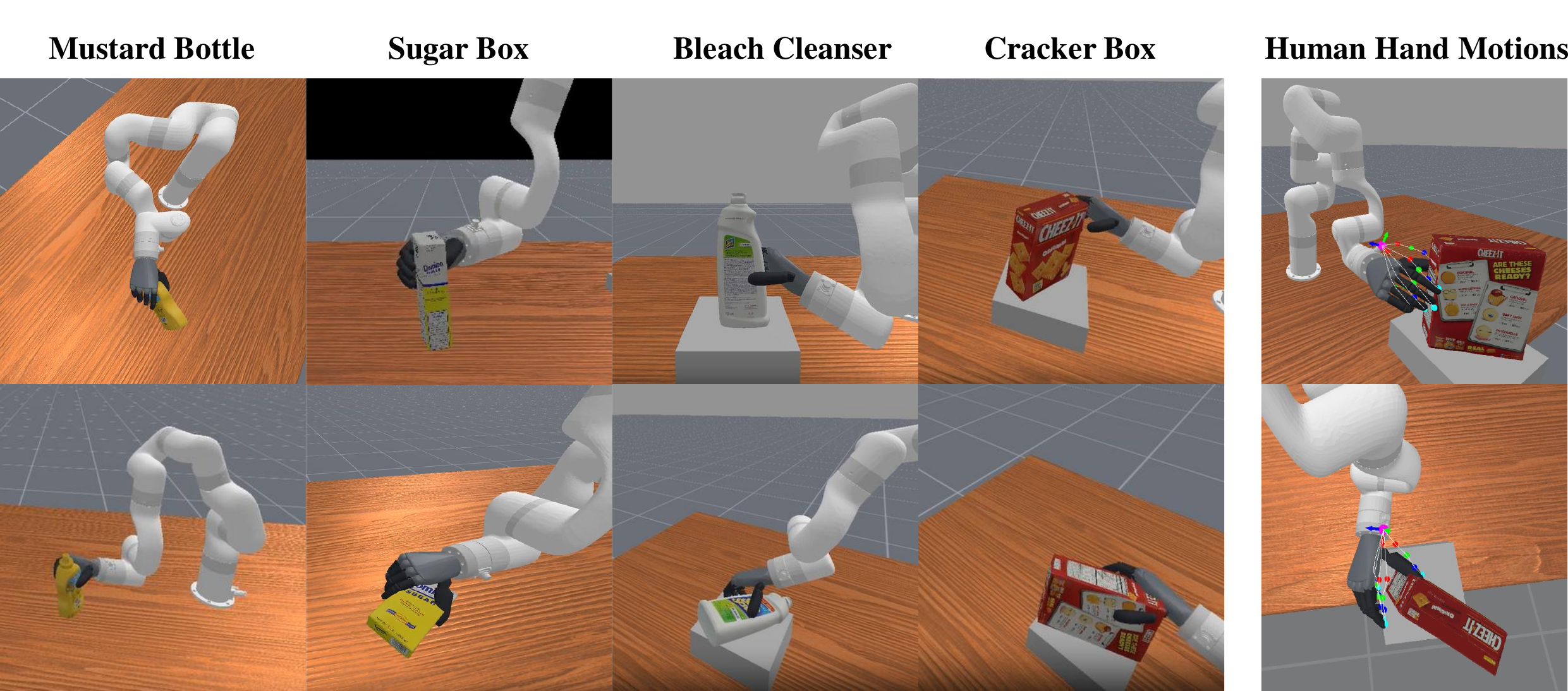}
    % \caption{Examples of robot Robot demonstrations for Ability Hand with XArm7 under different poses from selected DexYCB videos. Hand motion retargeting aligns fingertip positions but introduces imperfections due to the embodiment gap.} 
    \caption{Example robot demonstrations converted from human videos in the DexYCB dataset. The Ability Hand with xArm7 replicates human grasps under different object poses. }
    \vspace{-20pt}
    \label{fig:retargeting}
\end{figure}

Since human hand motion follows a minimum jerk trajectory \cite{flash1985coordination}, which reduces joint errors \cite{kyriakopoulos1988minimum} and minimizes mechanical wear \cite{merat1987introduction}, we apply post-processing to the joint position sequence $\{q_k\}_{k=0}^{T-1}$ using the model from \cite{todorov1998smoothness}, following \cite{qin2022dexmv}. Fig. \ref{fig:retargeting} illustrates examples of motion retargeting with an Ability hand mounted on an xArm7.  

% overview: 为了学会跟着demo抓取，训练分成两个阶段。（TODO:没有curriculum learning，直接学的消融实验，training curve...）第一阶段无rand，学会复现demo。第二阶段逐步加上rand。（TODO：不同的success bound的实验）

\subsection{Reward Function Designs} \label{sec:rl}

Our goal is to leverage human demonstrations to learn natural, human-like grasping strategies at high sample efficiency. We achieve this with a trajectory following reward that encourages the policy to stay close to the reference trajectory. Since human demonstrations may not perfectly translate to robotic execution due to differences in morphology and kinematics, the robot must balance guidance with exploration. To address this, we design a reward scheme that integrates the trajectory following reward with other rewards that measure task progression.

\textbf{Trajectory following reward.} 
The reward guides the robot hand toward the pre-grasp stage. Existing work \cite{chen2024vividex} enforces state-to-state alignment between the robot trajectory and the reference trajectory, which can be overly restrictive, as human demonstrations may contain suboptimal or infeasible motions. 
\short{Instead, agents should flexibly reach key states using their own strategies while leveraging demonstrations.}
%Instead, the agent should have the flexibility to reach key states using its own strategy while still leveraging human demonstrations.  
To achieve this, we propose a reward that measures the proximity of robot states and the reference trajectory, inspired by TR$^2$ \cite{tao2023abstract}. The reward at time step $t$ is defined as:
\begin{equation*}
    R^{(t)}_t = \left\{
    \begin{aligned}
        & r_{\text{dist}}(s_t, d_{k_{\text{max}}(s_t)}) \cdot \left( \eta + \beta \cdot k_{\text{max}}(s_t) \right), \\
        & \quad \text{if } k_{\text{max}}(s_t) > \max\limits_{0 \leq t^\prime < t} k_{\text{max}}(s_{t^\prime}), \\
        & 0, \quad \text{otherwise.}
    \end{aligned}
    \right.
    \label{eq:tr2}
\end{equation*}
where $k_{\text{max}}(s_t) = \max_{0 \leq k < T} \{ k \mid \text{dist}(s_t, d_k) < \epsilon \},$
% \begin{equation*}
%     \text{far}(s_t) = \max_{1 \leq k < T} \{ k \mid \text{dist}(s_t, d_k) < \epsilon \},
% \end{equation*}
tracks the furthest state in the reference trajectory that the policy has matched so far. Here, $T$ is the length of the reference trajectory, which may differ from the episode length $H$. The function $\text{dist}(s, d)$ measures the distance between state $s$ and a reference state $d$, defined as:
\begin{equation*}
    \text{dist}(s, d) = \sum_{i=1}^{5} \alpha_1 \Vert s^{i, \text{pos}} - d^{i, \text{pos}} \Vert_2 + 
    \alpha_2 \texttt{rad}(s^{i, \text{rot}}, d^{i, \text{rot}}).
    \label{eq:dist}
\end{equation*}
\short{Here, state $s^{i, *}, d^{i, *}$ denotes the relative pose of the $i$-th fingertip to the object, with $\text{pos}, \text{rot}$ indicating position and rotation, respectively. $r_{\text{dist}}(s, d)$ is defined as: $r_{\text{dist}}(s, d) = 1 - \tanh(\text{dist}(s, d)),$ mapping the distance to a bounded reward that encourages the policy to reach states closer to the reference. In this paper, we refer to our reward and that in \cite{chen2024vividex} as the trajectory following and trajectory mapping reward, respectively.}
%Here, state $s^{i, *}, d^{i, *}$ represents the relative pose of $i$-th fingertip to object pose, and $\text{pos}, \text{rot}$ denote the position and rotation of the pose separately. $r_{\text{dist}}(s, d)$ is defined as: $r_{\text{dist}}(s, d) = 1 - \tanh(\text{dist}(s, d)),$
%which maps the distance to a bounded reward that encourages the policy to visit states closer to the reference trajectory. In this paper, we refer to our proposed reward function and the reward in \cite{chen2024vividex} as trajectory following reward and trajectory mapping reward for distinction. 

We empirically set $\alpha_1=1, \alpha_2=0.03, \eta=30, \beta=0.2$, and $\epsilon=0.04$. The reward is assigned only when the policy makes progress toward reaching a further state in the reference trajectory. 
% Once the robot hand reaches the object, it remains relatively still, causing the furthest matched state index to jump to the last frame, leaving no useful guidance for manipulation. 
Once the robot hand reaches the object, the reward value remains zero as the furthest matched state index jumps to the last frame. As a result, the reward only guides the policy up to the pre-grasp stage.

\textbf{Contact reward.} A stable grasp requires sufficient contact between the hand and the object. Inspired by the concept of \textit{force closure} in grasp generation \cite{liu2021synthesizing, wang2023dexgraspnet}, a grasp is considered force-closure if there exist contact forces $\{f_i\}$ at contact points $\{x_i\}$ within the friction cones rooted at $x_i$ that can resist arbitrary external wrenches. While force-closure provides a robust measure of grasp stability, directly enforcing it is computationally complex. Instead, we design a heuristic contact reward ($R^{(c)}$) that is assigned when the thumb and at least two other fingers make contact with the object. This encourages grasps that approximate force-closure without requiring explicit force analysis.  

\textbf{Height reward.} To ensure task completion, we introduce a height reward ($R^{(h)}$), which is proportional to the object's height and provides an additional bonus when the object reaches the target height. This encourages the policy to complete the grasping task rather than stopping prematurely.  

The final reward at time $t$ is defined as:  
\begin{equation*}
    R_t = R^{(t)}_t + R^{(c)}_t(1 + R^{(h)}_t).
\end{equation*}

% Since our reward function consists only of a trajectory following reward, a contact reward, and a goal-related reward, it can be applied across various manipulation tasks, reducing the need for task-specific reward engineering.

\subsection{Curriculum Learning}\label{sec:cl}

\short{While the reward terms guide the learner to follow the reference and discover feasible robot trajectories, the policy may struggle to generalize to diverse initial object poses due to variation in demonstrations. To improve generalization, we use curriculum learning, gradually increasing task difficulty to expose the policy to a wider range of object configurations.}
% While the reward terms guide the learner to follow the reference trajectory and discover physically plausible robot trajectories to achieve the goal, the learned policy may struggle to generalize to varying initial object positions and orientations due to the diversity in demonstrations. To enhance generalization, we employ curriculum learning, progressively increasing task difficulty to expose the policy to a broader range of object configurations.

Initially, we use one fixed object's pose for learning so that the policy quickly discovers feasible grasp poses. We then gradually randomize the initial pose, defined as a tuple of position and rotation $(p^{\text{pos}}, p^{\text{rot}})$: $p^{\text{pos}} \sim \mathcal{U}(p_{\text{init}}^{\text{pos}} -\sigma P_{\text{max}}, p_{\text{init}}^{\text{pos}} +\sigma P_{\text{max}}), \quad p^{\text{rot}} \sim \mathcal{U}(p^{\text{rot}}_{\text{init}} - \sigma \Theta_{\text{max}},p^{\text{rot}}_{\text{init}} + \sigma \Theta_{\text{max}})$
% \begin{equation*}
%     p^{\text{pos}} \sim p_{\text{init}}^{\text{pos}} + \sigma \mathbb{P}, \quad p^{\text{rot}} \sim p^{\text{rot}}_{\text{init}} + \sigma \Theta
% \end{equation*}
where $(p_{\text{init}}^{\text{pos}}, p^{\text{rot}}_{\text{init}})$ is the original pose, $\mathcal{U}$ stands for uniform distribution, $P_{\text{max}}$ and $\Theta_{\text{max}}$ represent the maximum position displace in $x,y$ and the maximum rotation deviation around the $z$-axis, respectively. The noise factor $\sigma \in [0, 1]$ controls the degree of variation, which is initialized as $0$ and increased incrementally by $0.01$ until reaching $1$ after the performance exceeds a predefined success threshold $\zeta$. This ensures that the policy learns to generalize to diverse object configurations at a high sample efficiency. %\todo{refine} 

\subsection{Adaptive Skill Selection} \label{sec:skill}

Curriculum learning allows us to train a diverse set of grasping policies, $\Pi=\{\pi_i\}_{i=0}^{N-1}$, each corresponding to a different grasp pose. However, selecting the most appropriate policy based on human preference remains a challenge. To address this, we introduce a Vision-Language Model (VLM)-based agent as a decision-making module that retrieves the optimal skill from the skill library $\Pi$. Given a human instruction $I$ in natural language, the VLM processes the instruction, environment image, and skill descriptions to infer the most suitable grasping policy. The selected policy is then executed, ensuring alignment with human intent and the environmental context.

%Curriculum learning enables us to learn diverse object grasping posture, however, we still requires a agent to decide wOnce obtained a diverse grasping policy $\{\pi_i\}$, we . We introduce a Vision Language Model (VLM)-based agent to retrieve the most suitable skill. $\Psi$, natural language instruction $I$}
%The agent consists of two components: a skills inventory and a VLM-based brain. The skills inventory holds multiple policies, each corresponding to a different grasp pose. Each skill can be described using natural language. The VLM (Vision-Language Model) is used to decide which skill to apply. The agent receives two inputs: the environmental information and a human instruction. Initially, the VLM takes in the environment image, the human instruction, and descriptions of each skill. By analyzing the environment and the instruction, the VLM outputs the best skill to use, selecting the one that best fits the current environment and human preference. The agent then executes the selected policy to grasp the object.

\section{Experiments}

In this section, we investigate the following questions:

% \yulin{In experiments, we want to address the following questions: 
% \begin{itemize}
%     \item Does our proposed trajectory following reward term enables learning diverse human-like grasping poses and increase training efficiency? \todo{what about naturensss?} 
%     \item Why we choose a tr2 based reward instead of a direct segment mapping? 
%     \item the purpose of curriculum learning
%     \item adaptive skill selection
% \end{itemize}
% Then we showcase the ability of our VLM demos and conducts experiments on real world.

\begin{itemize}
    \item Does our framework enable sample-efficient acquisition of generalizable human-like grasp strategies?  
    \item Is our vision-language model-based skill selection adaptable to varying user preferences?  
    \item Can the policy learned in simulation transfer effectively to the real world?  
\end{itemize}  

\subsection{Evaluating the Learning Framwork}
\label{sec:eval_rl}

\subsubsection{\textbf{Evaluation Setup}}

% We first introduce our baselines and then present the empirical results in the subsequent subsections. Since our approach incorporates trajectory following for curriculum learning, the focus of the evaluation is primarily on the performance of these two components. We use three baselines for comparison:

\textbf{Task setup.} We focus on four objects from DexYCB\cite{chao2021dexycb} dataset: \mustard{}, \bleach{}, \cracker{}, and \sugar{}. For each object, we select two videos, each featuring a different initial position and grasp pose. %\figref{fig:retargeting} illustrates the converted robot demonstrations for two different initial postures of \bleach{}. 
To evaluate our learning framework for skill library construction, we train a separate policy for each object-pose combination. Performance is measured by the success rate (SR), defined as the proportion of successful object lifts that reach a sufficient height. Each experiment is repeated five times with independent trials, and we report the average performance metrics.

\textbf{Simulation environment.} We use a PSYONIC Ability Hand on an xArm 7, matching our real-world hardware setup. Training and testing are conducted in ManiSkill3 \cite{tao2024maniskill3}, a SAPIEN-powered \cite{xiang2020sapien} framework that enables realistic environments and fast GPU-parallelized training.

\textbf{Implementation.} \short{We adopt Proximal Policy Optimization (PPO) \cite{schulman2017proximal} for policy learning.
\update{The robot observes the full simulation state, including all joint angles, and the poses of the hand, fingertips, and objects, represented in the robot arm base frame. The policy outputs joint angle deltas, indicating real-time changes for actuation.}
For each object and grasp pose, we train the policy for 80M steps with a fixed initial pose, followed by another 120M steps using the proposed curriculum. During curriculum training, we randomize object position within $\pm 5$ cm in $xy$ and rotation within $\pm 30^\circ$ around the $z$-axis. A success is defined as lifting the object above 20 cm.
At test time, we evaluate the policy on varied initial object configurations, using the same randomization as in curriculum training.}
% We use Proximal Policy Optimization (PPO) \cite{schulman2017proximal} for policy learning. \update{The robot's observation consists of the full state information from the simulation environment, including all joint angles of the robot, as well as the positions of the hand, fingertips, and objects, expressed in the coordinate frame of the robot arm's base. The control policy outputs joint angle deltas, specifying the real-time variation in each joint angle for robot actuation.} For each object and one specific grasping pose, we first train the policy for 80M steps with a fixed initial object pose. After obtaining a stable policy, we continue training for an additional 120M steps with the proposed curriculum. Specifically, we apply randomization with an $xy$ translation range of $\pm 5$  cm and a rotation range of $\pm30^\circ$ around the $z$-axis at the end of curriculum training. A success is rewarded when the object's height is above $20$ cm.During testing, we evaluate the learned policies across different initial object configurations, using the same randomized strategy applied at the end of curriculum training.

\subsubsection{\textbf{Main Results}}

\short{We compare our method with the state-of-the-art ViViDex \cite{chen2024vividex}, which uses a trajectory mapping reward to aid RL training, followed by imitation learning on rollouts for better generalization. \update{The observation and action spaces are the same as in \ours{}.} For fair comparison, we trained both our reimplementation of ViViDex and \ours{} for 200M steps. \update{Our proposed reward design differs in several aspects. In the pre-grasp stage, we reward object approach and contact. These are retained during manipulation, with an added reward for object motion alignment. To address sim-demo length mismatch, we use progress-based frame selection. The object-approach reward extends ViViDex with rotation-aware distance, while the contact reward follows \secref{sec:rl}.}}
% We compare our method with the state-of-the-art, ViViDex \cite{chen2024vividex}, which uses a trajectory mapping reward to improve RL training, followed by imitation learning on rollouts for better generalization. \update{The observation and action spaces are identical to those used in \ours{}.} As the official implementation is unavailable, we implement the reward and perform training for 200M steps. \update{Following ViViDex, each demonstration is segmented into pre-grasp and manipulation stages. In the pre-grasp stage, rewards are given for object approach and contact; in the manipulation stage, an additional reward encourages alignment with object motion. Since simulation steps may not match demonstration length, we replace frame-by-frame matching with proportional frame selection based on episode progress. The object-approach reward follows ViViDex's design but incorporates rotational differences in the distance metric. The contact reward remains as defined in \secref{sec:rl}.}  
To ensure fairness, we use the same randomized initial object poses from our curriculum training for both training and testing. In addition, we implement a version of ViViDex with our proposed learning curriculum, which differs from \ours{} only in the reward function design, for a more controlled comparison.

Our results show that ViViDex struggles with our demonstration dataset, even with our proposed curriculum. This is likely due to its sensitivity to the morphological gap between the human hand and the robot. The Ability Hand’s smaller size requires greater precision in grasping, while the trajectory mapping reward enforces strict alignment with the imperfect reference trajectory extracted from human demonstration. This highlights the robustness of our reward function, which we further analyze in \secref{sec:map}.

\begin{table*}[htp]
    \centering
    \begin{tabular}{c|cc|cc|cc|cc|c}
    \toprule
         \multirow{2}{*}{SR ($\%$) $\uparrow$} & \multicolumn{2}{c|}{\mustard{}} & \multicolumn{2}{c|}{\bleach{}} & \multicolumn{2}{c|}{\cracker{}} & \multicolumn{2}{c|}{\sugar{}} & \texttt{Avg} \\ 
         & pose-1 & pose-2 & pose-1 & pose-2 & pose-1 & pose-2 & pose-1 & pose-2 &  \\
    \midrule
      ViViDex\cite{chen2024vividex} & $0.00$ & $0.02$ & $0.00$ & $0.00$ & $0.00$ & $0.39$ & $0.04$ & $0.00$ & $0.06$ \\ 
      ViViDex (w/ curriculum) & 0.00 & 0.00 & 0.00 & 0.00 & 0.00 & 11.74 & 0.08 & 0.00 & 1.48\\
      Ours & $\mathbf{61.45}$ & $\mathbf{84.38}$ & $\mathbf{98.89}$ & $\mathbf{99.55}$ & $\mathbf{76.56}$ & $\mathbf{94.24}$ & $\mathbf{67.34}$ & $\mathbf{98.61}$ & $\mathbf{85.12}$\\
    \bottomrule
    \end{tabular}
    \caption{Quantitative results in simulation. We evaluate our policies, each trained to learn different grasp poses from distinct human demonstrations, on objects with varying visual appearances, sizes, and physical properties.} \vspace{-20pt}
    \label{tab:main_result}
\end{table*}

\subsubsection{\textbf{Analysis of Trajectory Following vs. Trajectory Mapping Reward}}\label{sec:map} To compare trajectory following and trajectory mapping rewards, we implement a variant using trajectory mapping reward and conduct experiments on \sugar{} pose-2. We measure the sum of fingertip distances from the demonstration’s \textit{pre-grasp} frame and track object height.

\begin{figure}[t]
    \centering
    \begin{subfigure}[b]{1\linewidth}
    % \captionsetup[]{skip=0pt}
    \includegraphics[width=\linewidth]{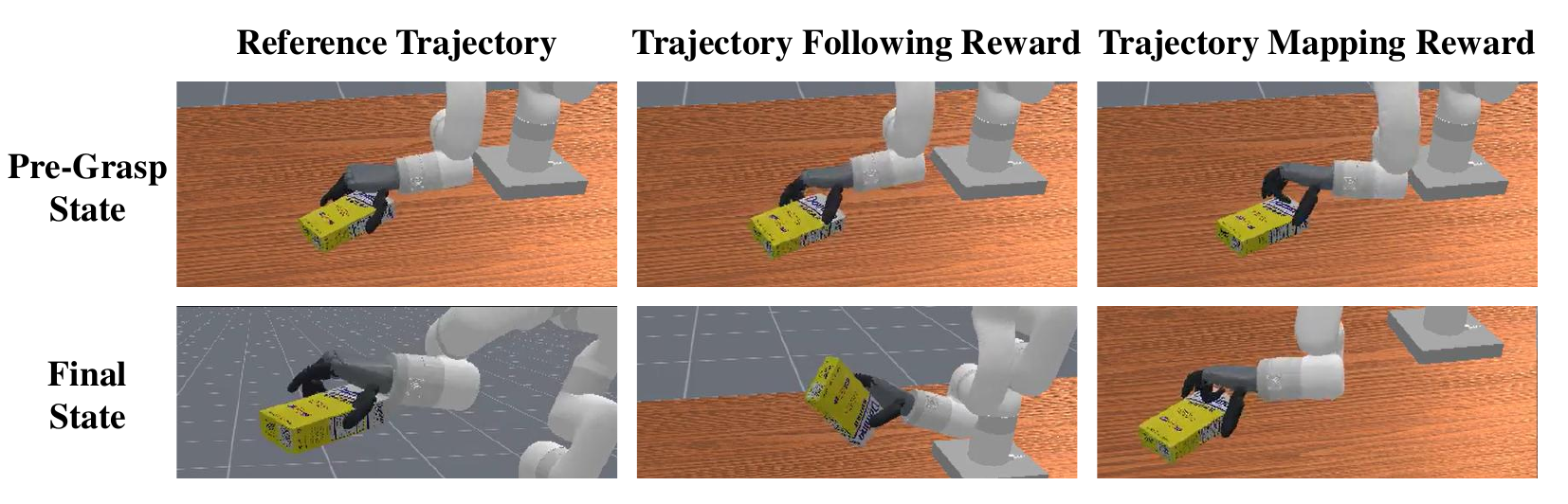}
    \vspace{-18pt}
    \caption{Illustration of pre-grasp and final grasp states.} \vspace{5pt}
    \end{subfigure}

    % \vspace{-10pt}
    \begin{subfigure}[b]{1\linewidth}
    \includegraphics[width=\linewidth]{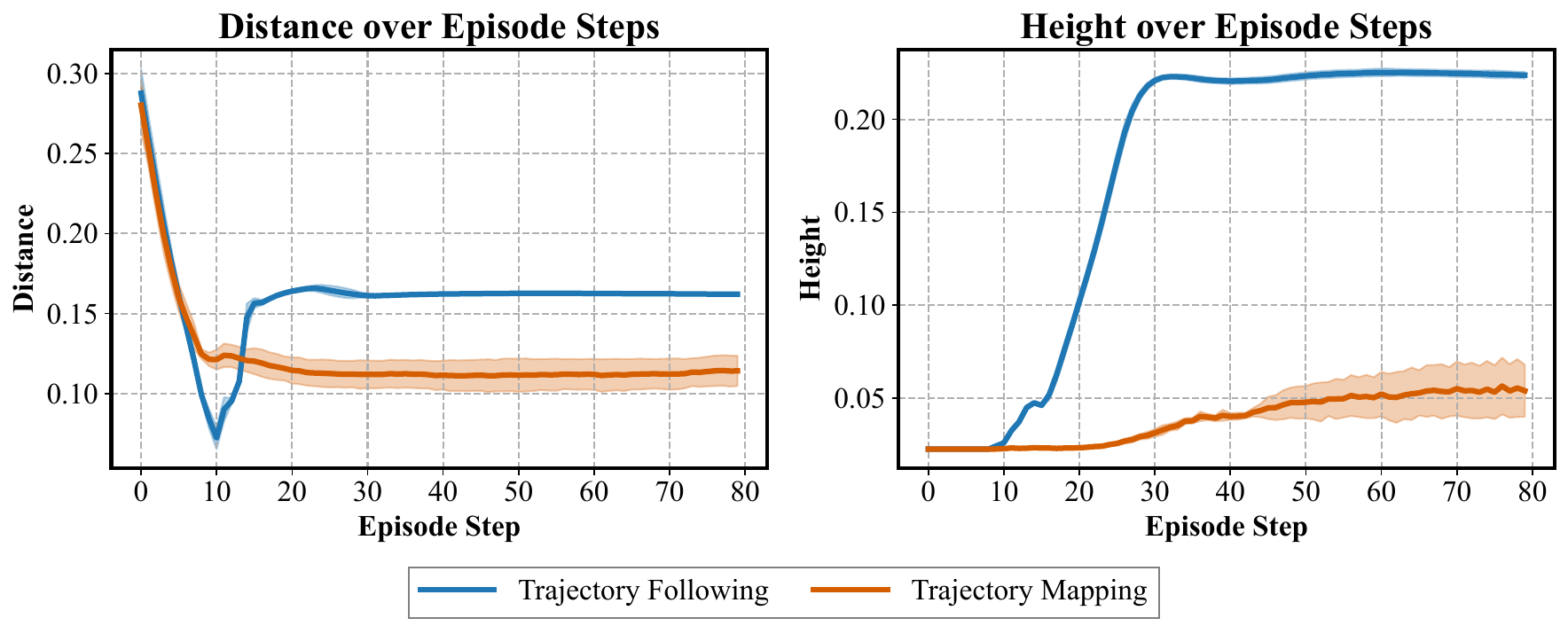}\vspace{-5pt}
    \caption{Fingertip distance and object height over episode steps.}
    \end{subfigure}\vspace{-2pt}
    \caption{Comparison between the proposed trajectory following reward and a standard trajectory mapping reward.} \vspace{-20pt}
    \label{fig:tf}
\end{figure}

As shown in \figref{fig:tf}, our reward function enables the robot to achieve a grasp pose closer to the demonstration before contact. However, after grasping, the fingertip distance increases, exceeding that of trajectory mapping. The success of our policy, despite this deviation, suggests that strict adherence to the reference trajectory is not always optimal. Human demonstrations are inherently imperfect for robots due to morphological differences, and motion retargeting introduces additional approximation errors. Unlike trajectory mapping, our reward function allows flexibility in execution, enabling the RL policy to refine the grasp for better effectiveness.

% \begin{figure*}[htp]
%     \includegraphics[width=\linewidth, height=3in]{example-image}
%     \caption{success rate curve in stage 1 compared with vividex/ no traj/ no grasp / ours}
%     \label{fig:main_result}
% \end{figure*}
\subsubsection{\textbf{Ablation Study of Reward Terms}} \short{We conduct ablation studies across eight grasping environments with fixed initial object poses. We compare our full pipeline with variations where each reward term is omitted. Without contact rewards, height rewards are given without contact checks. Without height rewards, we remove the height-proportional reward but retain the success reward when the object reaches the target height. The results are shown in Fig. \ref{tab:ablation}.}
%We conduct ablation studies across eight grasping environments with fixed initial object poses to evaluate the impact of three reward terms. We compare our full pipeline with variations where each reward term is omitted. Without contact rewards, height rewards are given without contact checks. Without height rewards, we remove the height-proportional reward but retain the success reward when the object reaches the target height. The results are shown in Fig. \ref{tab:ablation}.

The baseline without trajectory following reward shows near-zero success, failing in eight environments, indicating its crucial role in reducing the sample space for RL training. While the variant without contact reward performs well in three environments, it struggles in others, showing its importance in refining hand-object interactions despite imperfect demonstrations. Similarly, removing the height reward lowers success rates, highlighting its role in guiding the policy with goal-related information.

\begin{table*}[t]
    \centering
    \begin{tabular}{c|cc|cc|cc|cc|c}
    \toprule
      \multirow{2}{*}{SR ($\%$) $\uparrow$} & \multicolumn{2}{c|}{\mustard{}} & \multicolumn{2}{c|}{\bleach{}} & \multicolumn{2}{c|}{\cracker{}} & \multicolumn{2}{c|}{\sugar{}} & \texttt{Avg} \\
    & pose-1 & pose-2 & pose-1 & pose-2 & pose-1 & pose-2 & pose-1 & pose-2 &  \\
    \midrule
      w/o trajectory following & $0.00$ & $0.00$ & $0.00$ & $19.94$ & $0.00$ & $0.02$ & $0.00$ & $0.00$ & $2.50$ \\
      w/o contact reward & $59.80$ & $0.00$ & $0.00$ & $0.00$ & $40.00$ & $83.98$ & $\mathbf{79.69}$ & $39.94$ & $37.93$ \\
      w/o height reward & $\mathbf{79.94}$ & $\mathbf{79.92}$ & $27.50$ & $0.00$ & $60.00$ & $77.29$ & $39.80$ & $20.00$ & $48.06$ \\
      Ours & $79.84$ & $39.84$ & $\mathbf{39.90}$ & $\mathbf{39.92}$ & $\mathbf{79.94}$ & $\mathbf{96.48}$ & $79.63$ & $\mathbf{59.34}$ & $\mathbf{64.36}$ \\
    \bottomrule
    \end{tabular}
    \caption{Ablation study in simulation. We evaluate the impact of removing individual design components and demonstrate that our full pipeline achieves the best performance.} \vspace{-20pt}
    \label{tab:ablation}
\end{table*}

\subsubsection{\textbf{Effects of Curriculum Learning}}

To evaluate the impact of curriculum learning on training efficiency, we conduct experiments on \sugar{} tasks with two grasp poses.  We first train a policy for 80M steps using a fixed initial object pose. Training then continues for an additional 120M steps, during which we systematically vary the success threshold $\zeta$ to analyze its effect. Additionally, we compare with direct training, where the policy is trained from the beginning with the same level of object pose randomization as at the end of curriculum learning.

During the early stages of curriculum learning, the task success rate initially declines before stabilizing around the success threshold $\zeta$. As the success rate remains near $\zeta$, the noise level gradually increases. Once the noise level reaches its final value, $\sigma=1$, the task success rate surpasses $\zeta$. 

\short{We observe that a lower success threshold causes a faster rise in noise level. However, when comparing the steps at which different settings reach the same success rate after the noise level hits 1, a higher threshold yields greater training efficiency. This suggests that maintaining a high success rate while gradually introducing noise during curriculum learning offers better sample efficiency than rapidly increasing noise. Furthermore, direct training results in a significantly lower success rate, underscoring the effectiveness of curriculum learning in improving RL training efficiency.}
% We observe that a lower success threshold leads to a faster increase in noise level. However, when comparing the steps at which different settings achieve the same success rate after the noise level reaches 1, a higher success threshold results in greater training efficiency. This suggests that maintaining a high success rate while gradually introducing noise during curriculum learning yields better sample efficiency than rapidly increasing the noise level. Furthermore, direct training leads to a significantly lower success rate, underscoring the effectiveness of curriculum learning in improving the efficiency of RL training.

\begin{figure}[t]
    \centering
    \includegraphics[width=\linewidth]{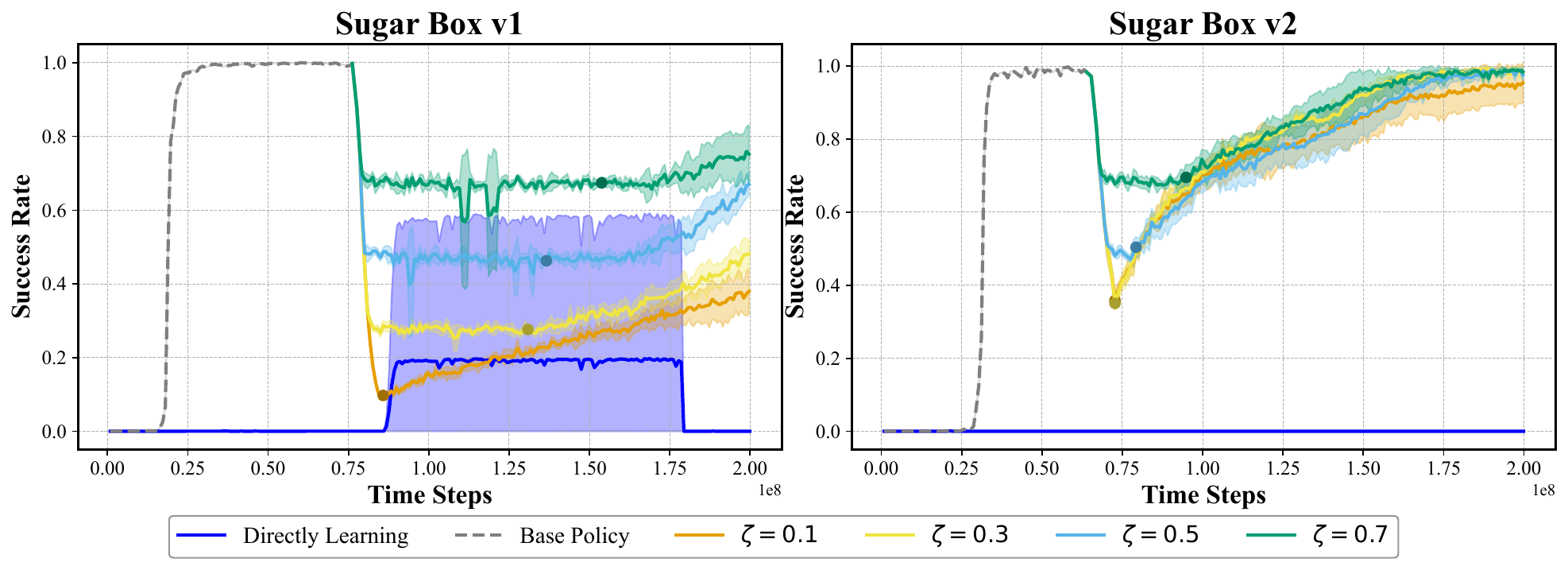}\vspace{-5pt}
    \caption{Comparison of sample efficiency between direct learning from fully random initial object pose and curriculum learning with different success thresholds ($\zeta$). The circular dot represents the average timestep at which $\sigma$ reaches 1. A suitable $\zeta$ leads to faster convergence and better performance. } \vspace{-5pt}
    \label{fig:analyze}
\end{figure}

\begin{table}
    \centering
    \begin{tabular}{c|ccc|cc}
    \toprule
     P(skill $|$ task) ($\%$) & T1 & T2 & T3 & T4 & T5 \\ 
    \midrule
    Skill 1 & $28$ & $100$ & $0$ & $0$ & $0$ \\
    Skill 2 & $72$ & $0$ & $100$ & $0$ & $0$ \\
    Skill 3 & $0$ & $0$ & $0$ & $100$ & $100$ \\  
    \bottomrule
    \end{tabular}
    \caption{The percentage of times each skill is selected for a given task. The Vision-Language Model (VLM) consistently assigns specific skills to tasks, indicating structured decision-making rather than random selection.} \vspace{-18pt}
    \label{tab:vlm1}
\end{table}

\begin{table}[t]
    \centering
    \begin{tabular}{c|ccc|cc}
    \toprule
    SR ($\%$) $\uparrow$ & T1 & T2 & T3 & T4 & T5 \\
    \midrule
    Random Selection & $44$ & $64$ & $64$ & $24$ & $52$ \\
    % VLM-based Selection & $0.96$ & $0.76$ & $1.00$ & $0.96$ & $0.96$ \\
    VLM-based Selection & $\mathbf{96}$ & $\mathbf{76}$ & $\mathbf{100}$ & $\mathbf{96}$ & $\mathbf{96}$ \\
    \bottomrule
    \end{tabular}
    \caption{Overall success rates of skill selection and execution. The table compares a random selection policy with a VLM-based policy, showing that VLM improves both skill selection and execution success.} \vspace{-18pt}
    \label{tab:vlm2}
\end{table}

\subsection{Evaluating Adaptive Skill Selection} 

% 设置：bleach物品，三个skill（分别是啥），5种任务（3立+2躺，2,3,5有语言偏好）。
% 1. 对比 T1 和 T2, T3，发现在没有语言偏好的时候机器人会抉择两个相似的skill，有语言偏好的时候会偏向语言指令。
% 2. 图像信息起很大作用，T5在有语言干扰的情况下仍然根据图像选择了正确的skill。

\subsubsection{\textbf{Evaluation Setup}}
% \quad 

\textbf{Task setup.}
% \todo{a figure?}
We evaluate the performance of the VLM-based agent for adaptive skill selection through a series of tasks designed to account for both the object's initial positions and human preference. Experiments are conducted on \bleach{} following the settings in Section \ref{sec:eval_rl}. The VLM operates based on two key principles: (i) ensuring successful object grasping and (ii) taking into account human preferences. 

Using our proposed RL training framework, we obtain three distinct skills (S1-S3): (i) Skill 1: Grasp the \textit{bottom} of a standing bottle and lift it. (ii) Skill 2: Grasp the \textit{upper middle} of a standing bottle and lift it. (iii) Skill 3: Grasp a lying bottle, rotate it upright, and lift it.

% We define five tasks varying the object's initial pose and human preference: (i) T1: Standing, no preference. (ii) T2: Standing, grasp \textit{bottom}. (iii) T3: Standing, grasp \textit{top}. (iv) T4: Lying, no preference. (v) T5: Lying, grasp \textit{bottom} (conflicting instruction). We note that in T3, \textit{top} is intentionally ambiguous, different from the \textit{upper middle} in S2 descriptions. In T5, the instruction contradicts the object’s pose, testing the VLM’s ability to resolve conflicts.

We define five tasks that vary based on the object's initial pose and the given human preference: (i) T1: Standing, no preference. (ii) T2: Standing, grasp \textit{bottom}. (iii) T3: Standing, grasp \textit{top}. (iv) T4: Lying, no preference. (v) T5: Lying, grasp \textit{bottom} (conflicting instruction). In T3, the term \textit{top} is intentionally ambiguous, as it differs from the \textit{upper middle} grasp described in S2. In T5, the instruction contradicts the object’s initial pose, evaluating the VLM’s ability to resolve conflicting grasping instructions.

\textbf{Implementation.} We use GPT-4 as the VLM. Given an environment image, human instructions, and descriptions of each skill, the VLM outputs the number of the selected skill along with a rationale for the choice in no more than three sentences.
% to the VLM, \slz{We designed a prompt to bootstrap the task, while also providing the environment image, human instructions, and descriptions of each skill to the VLM. We then asked the VLM to output the number of the selected skill, along with the reasoning for the choice in no more than three sentences.} 
Once the VLM selects a skill, the robot executes the corresponding policy to complete the task. Each task is evaluated 25 times with varying initial object poses. To assess the effectiveness of our method, we introduce a random selection baseline. This baseline randomly selects a skill from the skill library and executes it without considering the environment image or human instructions. %We use GPT-4 as the VLM brain. After the VLM selects a skill, the robot executes the corresponding policy to complete the task. For each task, we evaluate 25 times with different initial object pose. We report the skill selection and success rate for each task in Table \ref{tab:vlm}.

\subsubsection{\textbf{Results}} Skill selection results are reported in \tabref{tab:vlm1}. In T1, the VLM selects either S1 or S3. However, when a clear human preference is given (T2 and T3), the VLM reliably chooses the skill that best aligns with the request. In T4 and T5, the VLM prefers S3, as S1 and S2 are designed for a standing bottle. Even when instructed to grasp the bottom of a lying bottle (T5), the VLM selects S3 to ensure a feasible grasp. This shows that the VLM is able to consider both human intent and environmental context. 

\short{In terms of overall task success rate (\tabref{tab:vlm2}), random selection often fails to choose the appropriate skill. In contrast, VLM-based selection aligns with human preferences and achieves high task success.}
% In terms of overall task execution success rate (\tabref{tab:vlm2}), random selection often fails to select the appropriate skill. In contrast, VLM-based selection not only aligns well with human preferences but also achieves a high success rate in task execution.

\subsection{Real World Experiments}

\subsubsection{\textbf{Evaluation Setup}}
% \quad

\textbf{Task setup.} We validate the sim-to-real transfer of our policies on a real robot. 
% To demonstrate the effectiveness of our RL framework, we evaluate its performance on real-world dexterous grasping tasks. 
We focus on three objects with varying geometry, appearance, and mass distributions: \mustard{}, \bleach{}, and \cracker{}. For each object, we use the first grasp pose and apply a random $\pm 20^\circ$ rotation and a $\pm3$ cm position shift, repeating the experiment ten times.
% We utilize the DexYCB dataset, which contains human demonstration videos of hand-object interactions. In our experiments, we focus on five objects: a mustard bottle, a bleach bottle, a sugar box, a cracker box, and a power drill. For each object, we focus on 1 to 2 different grasp poses, as demonstrated by humans in the dataset.

\textbf{Hardware setup.} We use a PSYONIC Ability Hand mounted on an xArm 7 for real-world experiments. A RealSense D435i depth camera is calibrated using \cite{hong2024easyhec++} to capture RGB-D data for object pose estimation. %For simulation, we use the Maniskill platform, where we create a scene and load both the xArm and Ability Hand into the simulation environment.

\textbf{Implementation.} We use FoundationPose \cite{wen2024foundationpose} as our object pose estimator, which estimates object pose given object meshes, RGB-D observations, and object segmentation masks. At the start of each experiment, we provide a bounding box of the target object in the input image and use Segment Anything (SAM) \cite{kirillov2023segment} to extract its mask. After estimating the object pose, we initialize the object’s position in simulation to match the real-world setup and execute the policy for one episode of $80$ time steps. The resulting robot joint positions serve as tracking targets for the real robot. This approach reduces the sim-to-real gap compared to directly executing policy outputs, ensuring the real-world behavior aligns more closely with simulation \cite{liu2025dextrack}.

\begin{figure*}[t]
    \centering
    \begin{subfigure}[b]{0.49\linewidth}
    \includegraphics[width=\linewidth]{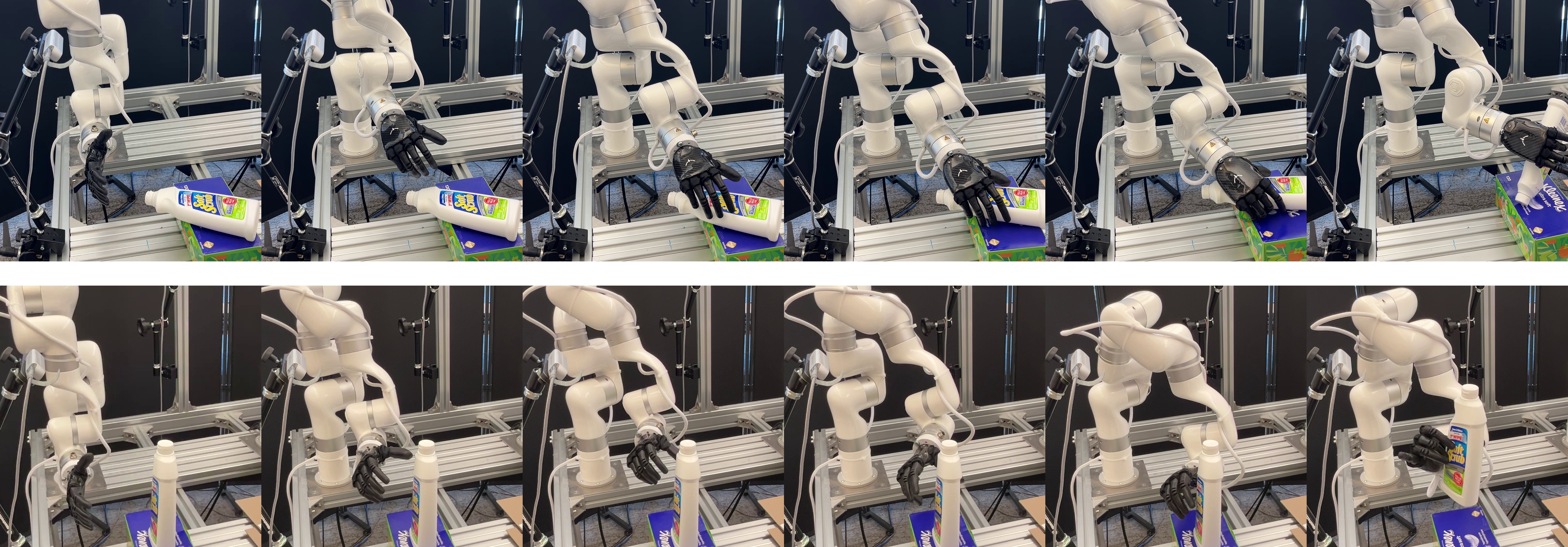}
    \vspace{-15pt}
    \caption{Grasping and lifting a bleach cleanser bottle.}
    \end{subfigure}
    \begin{subfigure}[b]{0.49\linewidth}
    \includegraphics[width=\linewidth]{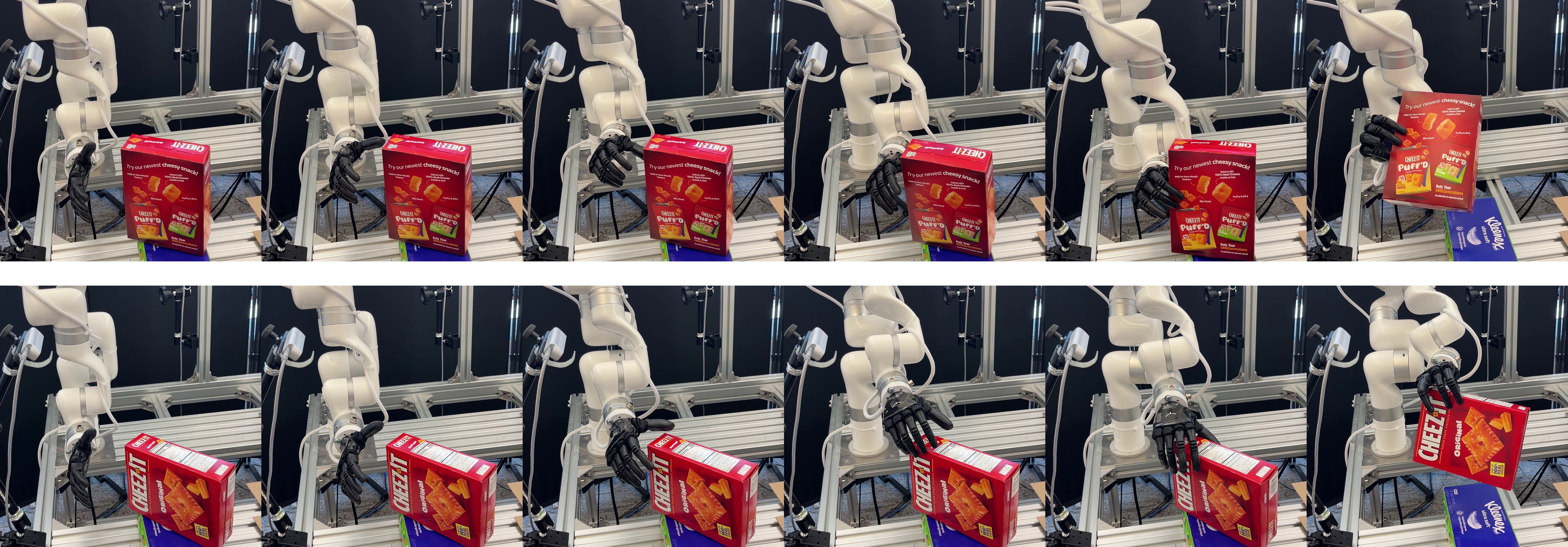}
    \vspace{-15pt}
    \caption{Grasping and lifting a cracker box.}
    \end{subfigure}\vspace{-5pt}
    \caption{Qualitative results illustrating the grasping process for the \mustard{} and \cracker{}. The images show the robot approaching, grasping, and lifting each object with two different grasp poses.} \vspace{-15pt}
    \label{fig:qualitative}
\end{figure*}

\setlength{\tabcolsep}{2pt}
\begin{table}[t]
    \centering
    \begin{tabular}{c|ccc}
    \toprule
    SR ($\%$) $\uparrow$ & \mustard{} & \bleach{} & \cracker{} \\ 
    \midrule
    ViViDex\cite{chen2024vividex} & $0$ & $0$ & $0$ \\
    Ours & $\mathbf{90}$ & $\mathbf{100}$ & $\mathbf{90}$ \\
    \bottomrule
    \end{tabular}
    \caption{Real-world manipulation results on objects with varying geometry, visual appearance, and physical properties. Our method outperforms the baseline by a large margin. } \vspace{-18pt}
    \label{tab:real}
\end{table}

\subsubsection{\textbf{Results}}

\short{As shown in \tabref{tab:real}, our method consistently achieves strong performance, maintaining at least a $90\%$ success rate across objects. In contrast, the baseline fails on all three objects, as expected given its near-zero success rate in simulation.
\figref{fig:qualitative} shows qualitative results, illustrating the robot grasping objects with different poses.}
% As shown in \tabref{tab:real}, our method consistently achieves strong performance, maintaining at least a $90\%$ success rate across objects. In contrast, the baseline fails on all three objects, which is expected given its near-zero success rate in simulation. \figref{fig:qualitative} presents qualitative results, illustrating the robot grasping objects with different poses.
\section{Conclusion \update{\& Limitations}}
We introduced \ours{}, a framework for efficiently learning and adaptively selecting dexterous grasping skills. By integrating human demonstrations, demonstration-guided reinforcement learning, curriculum learning, and VLM-based skill selection, \ours{} enables sample-efficient acquisition of diverse, human-like grasps. We demonstrate strong performance in both simulation and real-world experiments.

\update{Our approach has a few limitations. First, as the skill library scales to include more grasp and object types, retrieving the appropriate policy becomes more susceptible to hallucinations by VLMs. Structured knowledge representations (e.g., symbolic) may help encode skill affordances \cite{wu2025savor} and support more reliable information update and reasoning \cite{chen2024llm}. Second, while we test the model on simple, ambiguous instructions, its robustness to more complex or noisy language inputs remains to be explored. These challenges open promising directions for future work.}

\bibliographystyle{ieeetr}
\bibliography{references}

\end{document}